\documentclass[letterpaper, 10 pt, conference]{ieeeconf}
\IEEEoverridecommandlockouts
\usepackage{amsmath,amssymb}
\usepackage{graphicx}
\usepackage{booktabs}

\newcommand{\R}{\mathbb{R}}
\newcommand{\E}{\mathbb{E}}

\newcommand{\shB}{\mathrm{share}_B}

\title{\LARGE \bf
Diversity-Aware Experience Replay via Statistical Leverage Scores
}

\author{Anonymous authors}
\author{Shaifalee Saxena$^{1,2}$ and Alexander Scheinker$^{1}$%
\thanks{*This work was supported by Los Alamos National Laboratory}%
\thanks{$^{1}$Shaifalee Saxena and Alexander Scheinker are in the Instrumentation \& Controls Group at Los Alamos National Laboratory, Los Alamos, New Mexico 87545, USA. (Email: {\tt\small shaifalees@lanl.gov, ascheink@lanl.gov})
}
\thanks{$^{2}$Shaifalee Saxena is a PhD student at the University of New Mexico, Albuquerque, New Mexico, 87131, USA. (Email: {\tt\small shaifalisaxena@unm.edu})}
}

\begin{document}

\maketitle
\thispagestyle{empty}
\pagestyle{empty}

\begin{abstract}

Off-policy deep reinforcement learning trains on random samples from a replay buffer of past experience. In physical robot settings, that buffer is dominated by routine: thousands of near-identical transitions from normal operation, while the rare events that define the edges of the task are rarely sampled. We propose scoring every stored transition by its \emph{statistical leverage}, which measures how unusual it is relative to the buffer's own data, computed from the stored numbers alone, with no labels, rewards, or value network. We draw each mini-batch from a mixture of this score and uniform sampling. First, the natural way to build the score along the directions where the data varies most turns out to be blind when actions are stored as one-hot indicators, because indicator columns always win the variance ranking; we instead select the feature columns with the heaviest tails (largest kurtosis) where small far-away populations are revealed. Second, the importance weights that prioritized replay uses to keep gradients unbiased cancel the method exactly: with full correction, training matches uniform sampling, so we drop the correction and deliberately optimize the rebalanced objective. Third, when the rare situations of a log are relational, visible only in how a stretch of behavior relates to its goal, not in any single stored coordinate—per-transition scores are blind; applying the same tail rule to short trajectory windows described relative to their goal restores the signal, and on a three-joint arm that must push an object around an obstacle it raises the rare-task success from 0.49 to 0.69 against a labeled-oracle ceiling of 0.74.

\end{abstract}

\begin{figure*}[t]
\centering
\includegraphics[width=1.0\textwidth]{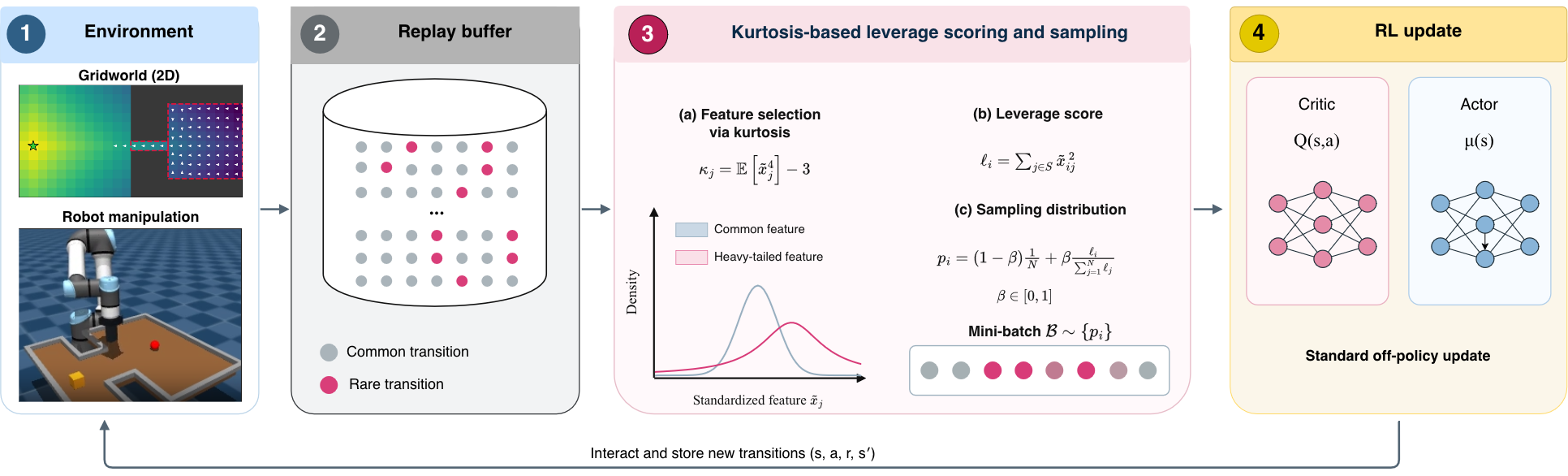}
\caption{Overview of the proposed kurtosis-based leverage-mixture replay framework.}
\label{fig:leverage}
\end{figure*}

\section{INTRODUCTION}

Deep reinforcement learning (RL) \cite{sutton2018reinforcement} has become a practical tool for synthesizing feedback controllers directly from interaction data. The modern off-policy template descends from the Deep Q-Network (DQN) \cite{mnih2015human}, which revived the classical idea of experience replay \cite{lin1992self}: transitions $(s,a,r,s')$ are stored in a large ring buffer, and mini-batches drawn from the buffer decorrelate consecutive updates and allow each interaction to be reused many times. Deep Deterministic Policy Gradient (DDPG) \cite{lillicrap2015continuous}, followed by Twin Delayed DDPG (TD3) \cite{fujimoto2018td3} and Soft Actor-Critic (SAC) \cite{haarnoja2018sac} carried this template to continuous control, where they remain the workhorse algorithms of robot learning \cite{ibarz2021how}. Sample efficiency has been improved mainly by extracting more gradient signal from the same stored data: ensembling and high update-to-data ratios \cite{chen2021redq}, normalization and network scaling \cite{bhatt2024crossq}, and learned state--action representations with prioritized replay \cite{fujimoto2023td7}. These advances support real-robot manipulation pipelines that train in minutes to hours \cite{ball2023rlpd,luo2024serl} and autonomous control of large scientific instruments, including the tuning of particle accelerators \cite{saxena2025improved,saxena2026mahalanobis}. Despite this progress, uniform sampling remains the default replay rule in standard off-policy RL.

Uniform sampling implicitly assumes that every stored transition is equally worth revisiting, and this assumption degrades precisely as learning succeeds. A partially converged policy fills the buffer with thousands of near-identical transitions, such as the gripper hovering over the object in almost the same pose, or the beam sitting at a near-optimal setpoint with very low beam loss. At the same time, rare transitions that delineate the boundaries of the task (unusual contacts, failed grasps, off-nominal beam excursions) are sampled with vanishing frequency and eventually overwritten. In this work, we make the buffer's own geometry the sampling signal. We view the buffer as a tall data matrix, one standardized transition per row, and score each transition by its \emph{statistical leverage} \cite{drineas2012leverage}: the squared, whitened length of its projection onto a chosen score subspace, equivalently a (truncated) Mahalanobis distance from the buffer's own empirical distribution. Mini-batches are drawn from a mixture placing weight $\beta$ on the normalized leverage scores and $1-\beta$ on the uniform distribution. Because $N \gg d$, the scores can be refreshed efficiently with buffer-wide statistics and a vectorized pass over the stored transitions.
Our contributions are: \textbf{(1)} a leverage-mixture replay rule using excess-kurtosis column selection to avoid variance-subspace blindness under one-hot actions, and omitting importance-sampling correction, which we show cancels the rebalancing exactly; \textbf{(2)} exact, numerically verified sampling-share laws on Gaussian-mixture buffers, including a rarity-independent amplification ceiling and a multi-cluster mass law; \textbf{(3)} offline Q-learning experiments against exact \(Q^*\), demonstrating improved learning in an underrepresented region; and \textbf{(4)} a window-level extension for relational rare behaviors, where goal-relative trajectory scoring on a three-joint arm trained with TD3+BC improves over transition-level replay and approaches a labeled oracle.

\begin{figure}[t]
\centering
\includegraphics[width=\columnwidth]{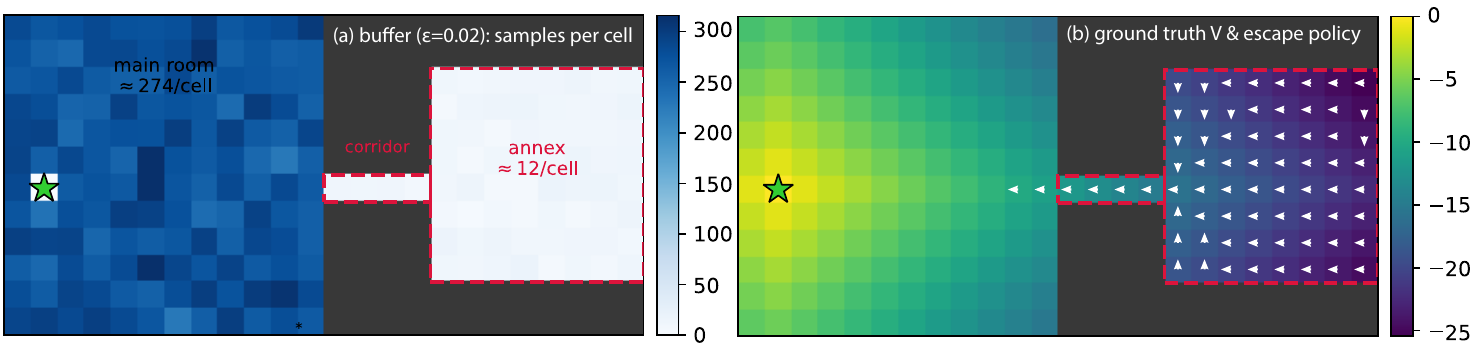}
\caption{Two-room gridworld. \textbf{(a)} The logged buffer at
$\varepsilon=0.05$: routine operation covers the main room ($\approx275$ samples/cell) while the corridor and annex region B, dashed, receive $\approx12$. \textbf{(b)} Ground-truth $Q^{*}$ with the optimal escape policy: the annex's value structure follows its own geometry and cannot be inferred from region-A data.}
\label{fig:grid_world}
\end{figure}
\section{RELATED WORK}

Non-uniform replay is not a new idea, and its adoption is strikingly uneven. Prioritized Experience Replay (PER) \cite{schaul2015prioritized}, which samples transitions in proportion to their temporal-difference (TD) error, is a standard ingredient of value-based and distributed agents.
Yet in the DDPG/TD3/SAC lineage that dominates continuous control, the de facto standard remains plain uniform sampling: the original algorithms use it \cite{lillicrap2015continuous,fujimoto2018td3,haarnoja2018sac}, and it is the default in many implementations.
Where modern robot-learning systems do bias the buffer, they use simple fixed composition rules rather than per-transition priorities: symmetric 50/50 sampling of online and offline data in RLPD and SERL \cite{ball2023rlpd,luo2024serl}.

Empirically, PER's gains transfer inconsistently from Atari to continuous control, where studies find it often matches or underperforms uniform sampling \cite{fujimoto2020equivalence,lahire2022laber}. Theoretically, prioritized sampling combined with its importance-sampling correction is largely equivalent to uniform sampling under a reshaped loss, so much of its effect can be replicated (or cancelled) by loss design \cite{fujimoto2020equivalence}; priorities are computed only when a transition was last visited and therefore go stale; the scheme introduces sensitive hyperparameters; and TD-error-based scores systematically chase noise and outliers in stochastic environments \cite{schaul2015prioritized,lahire2022laber,pan2022per}. Broader studies of replay find that buffer capacity, replay ratio, and retention policy often matter as much as the sampling rule itself \cite{fedus2020revisiting,zhang2017deeper,debruin2018experience}. When outlier sensitivity is repaired, as in the loss-adjusted prioritization used in TD7 \cite{fujimoto2020equivalence,fujimoto2023td7}, prioritization becomes beneficial again in continuous control, suggesting that the obstacle is not non-uniform sampling per se but the fragility of error-based scores. This motivates a complementary signal that does not depend on the current critic at all.

A smaller body of work targets buffer \emph{coverage} directly. Experience-selection studies show that which transitions are retained and replayed can matter as much as how many \cite{debruin2018experience}; determinantal point processes (DPPs) have been used to select diverse trajectories and goals within hindsight replay \cite{dai2021dtgsh} and to prioritize diverse transitions in high-dimensional environments \cite{zhao2025eder}. The recurring obstacle in this line is computational: DPP kernels over $N$ stored transitions are expensive, and much of the engineering effort in these works goes into approximating them. Our proposal can be read as the cheap linear-algebra limit of this idea: rank-$k$ leverage scores are exactly the single-point marginals of a projection DPP, computable for the entire buffer in $O(Ndk)$ per refresh and maintainable incrementally at $O(d^{2})$ per transition. To the best of our knowledge, leverage-score sampling has not previously been studied as a replay-sampling rule for RL.

Recent robot learning has increasingly shifted toward large-scale imitation and vision-language-action models \cite{black2024pi0}. These trends reduce the role of replay in some settings, but do not eliminate the regime considered here: when physical interaction is expensive, sample-efficient off-policy learning remains important \cite{ball2023rlpd,luo2024serl}. Recent work such as FastTD3 further shows that replay-based off-policy learning can remain competitive even at large scale and transfer to real hardware \cite{seo2025fasttd3}, while emerging VLA systems are themselves beginning to incorporate reinforcement-learning post-training from autonomously collected experience \cite{guo2025improving}. Thus, the replay buffer is not disappearing; the question is how stored experience should be selected and reused.

\section{LEVERAGE-MIXTURE REPLAY}
\label{sec:method}

\subsection{Buffer geometry and leverage scores}
Each stored transition contributes a feature row $x_i\in\R^{d}$; a buffer of size $N$ is the matrix $X\in\R^{N\times d}$. Which columns enter $x_i$ is itself a design decision that validation forced us to make explicit (Sec.~\ref{sec:columns}); by default $x_i$ contains the continuous state features. Columns are standardized,
\begin{equation}
\bar{x}_{ij} = \frac{x_{ij}-\mu_j}{c_j+\epsilon_0},
\qquad \epsilon_0\approx 10^{-8},
\label{eq:standardize}
\end{equation}
with $\mu_j, c_j$ the buffer mean and standard deviation of column $j$. The thin SVD $\bar{X}=U\Sigma V^{\top}$ gives principal directions $v_j$, PCA scores $T=\bar{X}V$, and covariance eigenvalues $\lambda_j=\sigma_j^{2}/(N-1)$ of $C=\tfrac{1}{N-1}\bar X^{\top}\bar X$. For a score subspace spanned by an index set $\mathcal{K}$ of directions, the leverage score of transition $i$ is
\begin{equation}
\ell_i \;=\; \frac{1}{N-1}\sum_{j\in\mathcal{K}}\frac{\big(\bar{x}_i^{\top}v_j\big)^{2}}{\lambda_j},
\qquad
0\le\ell_i\le 1,\quad \sum_{i=1}^{N}\ell_i=|\mathcal{K}|.
\label{eq:leverage}
\end{equation}
With $\mathcal{K}=\{1,\dots,d\}$ and invertible $C$ this is exactly the squared Mahalanobis distance of $x_i$ from the buffer's own mean and covariance, divided by $N-1$:
\begin{equation}
\ell_i=\tfrac{1}{N-1}\,\bar x_i^{\top}C^{-1}\bar x_i
=\tfrac{1}{N-1}\,(x_i-\mu)^{\top}S^{-1}(x_i-\mu),
\label{eq:mahal}
\end{equation}
where $S$ is the raw-coordinate covariance; the standardization cancels. Equivalently, up to constants $\ell_i$ is the negative log-density of transition $i$ under a Gaussian fit to the buffer: high leverage means precisely ``improbable under the buffer.'' Leverage scores are also the single-point inclusion marginals of the projection DPP with kernel $K=U_{\mathcal K}U_{\mathcal K}^{\top}$, so this is the cheap, independent-sampling limit of DPP-style diverse batch selection. Because $N\gg d$, we never form the $N\times N$ side: one $O(Nd^{2})$ pass (or $O(d^{2})$ incremental moments) plus an $O(d^{3})$ eigendecomposition, then all $N$ scores in one matrix multiply.

\subsection{Which columns to score, and how to pick the subspace}
\label{sec:columns}
Version 1 of this method scored $[\,s\,|\,a\,]$ rows and chose $\mathcal{K}$ as the top-$k$ principal directions with the smallest $k$ capturing $90\%$ of variance. Validation shows both choices fail.

\emph{Categorical blindness.} Suppose the action enters as a $K$-level one-hot block. Then $a$ takes $K$ equally likely values encoded as indicators $\mathbf{1}_j$. Each standardized column has unit variance, and for $j\ne k$, $\mathrm{Cov}(\mathbf{1}_j,\mathbf{1}_k)=-1/K^{2}$ gives correlation $-\tfrac{1}{K-1}$. The block correlation matrix is therefore $R=\tfrac{K}{K-1}I-\tfrac{1}{K-1}J$, with eigenvalue $0$ on the all-ones vector and $K/(K-1)$ with multiplicity $K-1$ on its complement. Since every standardized \emph{continuous} column has variance exactly $1$, the one-hot contrasts always head the ranking ($4/3$ for $K{=}4$, $3/2$ for $K{=}3$). Rare-mode separation is therefore hidden \emph{inside} that unit budget. Variance ranking therefore always places the action contrasts first: in our gridworld ($K{=}4$) the measured $\phi(s,a)$ spectrum is $(1.334,1.333,1.332,1.014,0.986,0)$.

\emph{Low-variance separating directions.} If the rare mode separates along a direction whose within-mode variance is small, truncated leverage is blind until $k$ reaches that direction's variance rank $r$ and diluted beyond it (Sec.~\ref{sec:analysis}); the useful subspace is the \emph{smallest} one containing the separating directions, which variance ordering does not target.

\emph{The amended rule.} We select \emph{columns}, not eigendirections, by a tail statistic: the excess kurtosis $\kappa_j=\E[\bar x_{\cdot j}^{4}]-3$ of each standardized continuous column, keeping the top-$m$ columns ($m{=}1$ by default) and computing \eqref{eq:leverage} within them. For $m{=}1$ the selected direction is the coordinate axis $e_{j^{*}}$, where $j^{*}=\arg\max_{j}\{\kappa_{j} \}$, rather than an eigenvector, so the projection in \eqref{eq:leverage} is the standardized entry itself, $\bar{x}_i^{\top}e_{j^{*}}=\bar{x}_{ij^{*}}$, whose normalizing variance $e_{j^{*}}^{\top}C\,e_{j^{*}}=1$ by standardization (up to the immaterial $N/(N-1)$ convention factor), giving
\begin{equation}
\ell_i \;=\; \frac{\bar{x}_{ij^{*}}^{2}}{N-1}\;\propto\;\bar{x}_{ij^{*}}^{2}.
\label{eq:tailscore}
\end{equation}
No eigendecomposition is required. The normalization
$\sum_{i=1}^{N}\ell_i=|\mathcal{K}|=m$ of \eqref{eq:leverage} is preserved, and because the mixture \eqref{eq:mixture} uses only the normalized scores $\ell_i/\sum_k\ell_k$, the prefactor is immaterial: a transition's score is simply the squared standardized value of its heaviest-tailed feature.

A column carrying a rare mass fraction $\varepsilon$ at standardized distance $z$ gains $\approx\varepsilon z^{4}$ of kurtosis, while uniform or Gaussian background columns are platykurtic; the statistic is the classical projection-pursuit signal for interesting directions \cite{huber1985projection}. Selecting fixed axes avoids the eigenvector degeneracy that per-column standardization induces (uncorrelated standardized columns give $C\approx I$, so principal directions rotate arbitrarily across seeds), and is immune to the one-hot pathology. The selection is not free of failure modes of its own---eigen-direction moments can be captured by skewed categorical contrasts, which is why we restrict to continuous columns---and full state-block Mahalanobis ($\mathcal{K}$ = all state directions) is the selection-free fallback, in which case \eqref{eq:leverage} is evaluated over all state directions and
collapses to the Mahalanobis form \eqref{eq:mahal}; this variant is selection-free and never blind, but pays the dilution cost of
\eqref{eq:dilution} with $k=d_s$. Reward is excluded from scoring: including $r$ re-introduces outcome-chasing, the very failure of error-based priorities the geometric signal is meant to complement; categorical action encodings are excluded whenever the logged action marginal is region-independent, since they then contribute only dilution and the pathology above.

\section{ANALYTIC GAUSSIAN MIXTURE STUDY}
\label{sec:analysis}

\begin{figure*}[t]
\centering
\includegraphics[width=\textwidth]{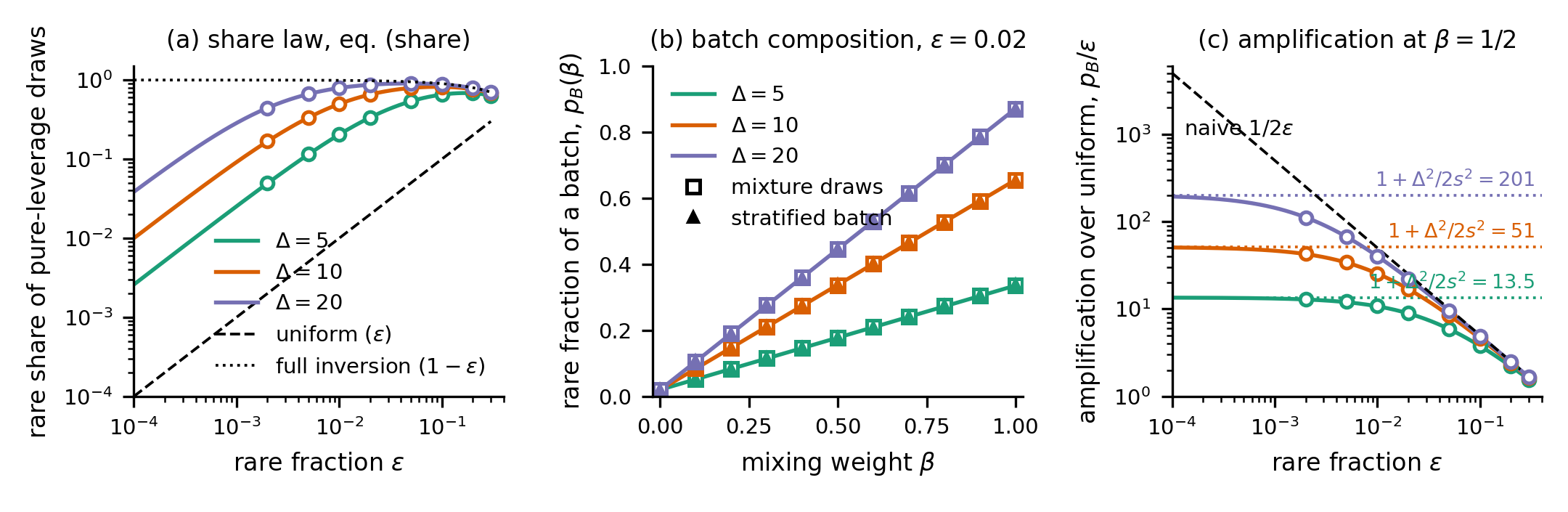}
\caption{Sampling-share laws on two-cluster Gaussian buffers (\(N=2\times10^5\), \(d=20\), \(s_A=s_B=s=1\)): theory (lines) vs. simulation (markers). (a) Rare share under pure-leverage sampling; dashed: uniform sampling \((\epsilon)\), dotted: full-inversion limit \((1-\epsilon)\). (b) Rare batch fraction versus mixing weight \(\beta\) at \(\epsilon=0.02\), for mixture draws (squares) and stratified batches (triangles). (c) Amplification over uniform sampling at \(\beta=1/2\); dotted: theoretical ceiling \(1+\Delta^2/(2s^2)\), dashed: naive \(1/(2\epsilon)\).}
\label{fig:theory}
\end{figure*}

\begin{figure}[t]
\centering
\includegraphics[width=\columnwidth]{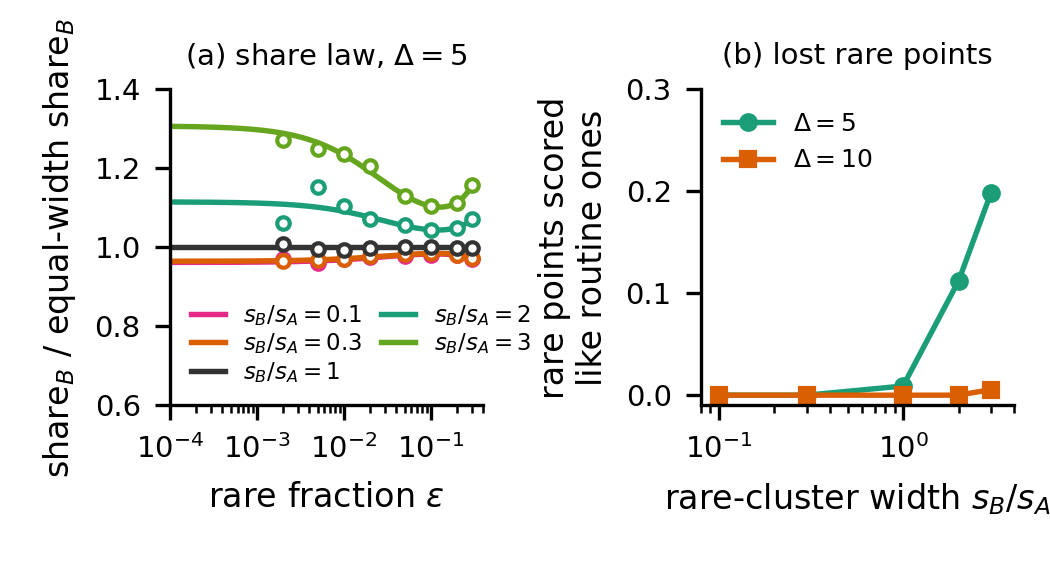}
\caption{Effect of rare cluster width ($s_A{=}1$). \textbf{(a)} Rare share of pure-leverage draws at $\Delta{=}5$ for $s_B/s_A$ from $0.1$ to $3$, divided by the equal-width value: lines \eqref{eq:share}. \textbf{(b)} Fraction of rare points whose score falls below the $99$th percentile of the common cluster's scores ($\varepsilon{=}0.01$).
}
\label{fig:width}
\end{figure}

We first study the sampling rule analytically in an idealized scenario where the standardized buffer $\bar X$ of \eqref{eq:standardize} is a mixture of two Gaussian clusters: a \emph{common} cluster A holding a fraction $1-\varepsilon$ of the transitions and a \emph{rare} cluster B holding the remaining fraction $\varepsilon$. The two clusters differ along a single direction, described by a unit vector $u$ and their means are a distance $\Delta$ from each other giving the mixture $(1-\varepsilon)\mathcal{N}(\mu_A,s_A^{2})+\varepsilon\,\mathcal{N}(\mu_B,s_B^{2})$. By projecting each transition onto it, $t_i=u^{\top}\bar x_i$, we get one number per transition, its coordinate along $u$. Along $u$ the common cluster scatters around its center with standard deviation $s_A$ and the rare cluster around its own center with standard deviation $s_B$. The buffer is centered, so the weighted average of the two centers is zero: the common center sits at $-\varepsilon\Delta$ and the rare center at $(1-\varepsilon)\Delta$. The total variance of $t_i$ is the average within-cluster variance $\bar s^{2}=(1-\varepsilon)s_A^{2}+\varepsilon s_B^{2}$ plus the extra variance created by the two centers being apart,
\begin{equation}
\lambda=\bar s^{2}+(1-\varepsilon)(\varepsilon\Delta)^{2}+\varepsilon\big((1-\varepsilon)\Delta\big)^{2}
       =\bar s^{2}+\varepsilon(1-\varepsilon)\Delta^{2} .
\label{eq:lambda}
\end{equation}
With $\mathcal{K}=\{u\}$, the leverage score \eqref{eq:leverage} is $\ell_i=t_i^{2}/[(N-1)\lambda]$: the squared coordinate along $u$, normalized so that the scores sum to one over the buffer. Under pure leverage sampling ($\beta=1$ in \eqref{eq:mixture}) the probability of drawing a rare transition is therefore the total score of cluster B divided by the total score of the buffer. The average of $t_i^{2}$ inside a cluster is its variance plus its squared center, so the average score of a rare point and of a common point, relative to the buffer-wide average score of $1$, are
\begin{equation}
m_B=\frac{s_B^{2}+(1-\varepsilon)^{2}\Delta^{2}}{\lambda},\qquad
m_A=\frac{s_A^{2}+\varepsilon^{2}\Delta^{2}}{\lambda},
\label{eq:mAmB}
\end{equation}
with $(1-\varepsilon)m_A+\varepsilon m_B=1$. Cluster B holds a fraction $\varepsilon$ of the points, each worth $m_B$ average points, so its share of pure-leverage draws is $\varepsilon m_B/[\varepsilon m_B+(1-\varepsilon)m_A]=\varepsilon m_B$:
\begin{equation}
\shB=\frac{\varepsilon\big(s_B^{2}+(1-\varepsilon)^{2}\Delta^{2}\big)}{(1-\varepsilon)s_A^{2}+\varepsilon s_B^{2}+\varepsilon(1-\varepsilon)\Delta^{2}} ,
\label{eq:share}
\end{equation}
verified to $0.003$ absolute error across $\varepsilon\in[0.002,0.3]$, $\Delta\in\{5,10,20\}$ for equal widths (Fig.~\ref{fig:theory}a) and to $0.006$ for $s_B/s_A\in[0.1,3]$ (Fig.~\ref{fig:width}a). Everything in \eqref{eq:share} is governed by one dimensionless number, $\varepsilon(1-\varepsilon)\Delta^{2}/\bar s^{2}$: the variance ``bump'' the rare cluster adds along $u$, relative to the within-cluster variance. If $\Delta=0$ the clusters are indistinguishable and $\shB=\varepsilon$, i.e.\ uniform sampling. If the bump is large, $\varepsilon\Delta^{2}\gg\bar s^{2}$, then $\shB\to1-\varepsilon$: the rare cluster receives the share the common cluster used to have, a full inversion of the buffer's composition; for a $1\%$ event this needs a separation of well over ten standard deviations. If the bump is small, $\varepsilon\Delta^{2}\ll s_A^{2}$, then $\shB\approx\varepsilon(\Delta^{2}+s_B^{2})/s_A^{2}$.

The rare cluster's own width plays only a minor role, as seen in \eqref{eq:mAmB}: a rare point's score is its squared distance from the buffer mean, which is $\Delta^{2}$ from the offset plus $s_B^{2}$ from the cluster's internal scatter, measured against a total variance that is set by the common cluster's width $s_A$ (and by the bump). Therefore $s_B$ only enters additively next to $\Delta^{2}$. At $\Delta=5$ and $\varepsilon=0.01$ a tight rare cluster ($s_B\le0.3\,s_A$) receives $3\%$ less share than an equal-width one and a diffuse cluster ($s_B=3\,s_A$) $24\%$ more; at $\Delta=10$ all widths are within $4\%$ of each other (Fig.~\ref{fig:width}a). A tight rare cluster is, if anything, found more cleanly: all of its points sit at the same distance from the mean, so none of them is scored like a routine transition, whereas for $s_B=3\,s_A$ at $\Delta=5$ one fifth of the rare points score below the $99$th percentile of the common cluster (Fig.~\ref{fig:width}b). Leverage looks for distance from the bulk, not for local density, so a compact rare mode cannot hide from it the way it can from a density- or nearest-neighbor-based novelty score. The column-selection rule behaves the same way: in these sweeps the separating column has the largest excess kurtosis of all $20$ columns for every $\varepsilon\le0.1$ and every width tested, so the rule of Sec.~\ref{sec:columns} finds it without labels; between $\varepsilon=0.1$ and $0.3$ the column turns from heavy-tailed into bimodal and its excess kurtosis becomes negative (in the tight, far limit the column is a two-point distribution whose excess kurtosis crosses zero at $\varepsilon\approx0.21$; wider rare clusters cross later), so the rule stops selecting it, which is consistent with its purpose as a rare-mode detector. Under the mixture rule \eqref{eq:mixture} a fraction $1-\beta$ of each batch is drawn uniformly and $\beta$ by leverage, so the expected rare fraction of a batch is
\begin{equation}
p_B(\beta)=(1-\beta)\,\varepsilon+\beta\,\shB ,
\label{eq:batchshare}
\end{equation}
and $p_B(\beta)/\varepsilon$ is rare cluster \emph{amplification} over uniform.

\begin{figure*}[t]
\centering
\includegraphics[width=\textwidth]{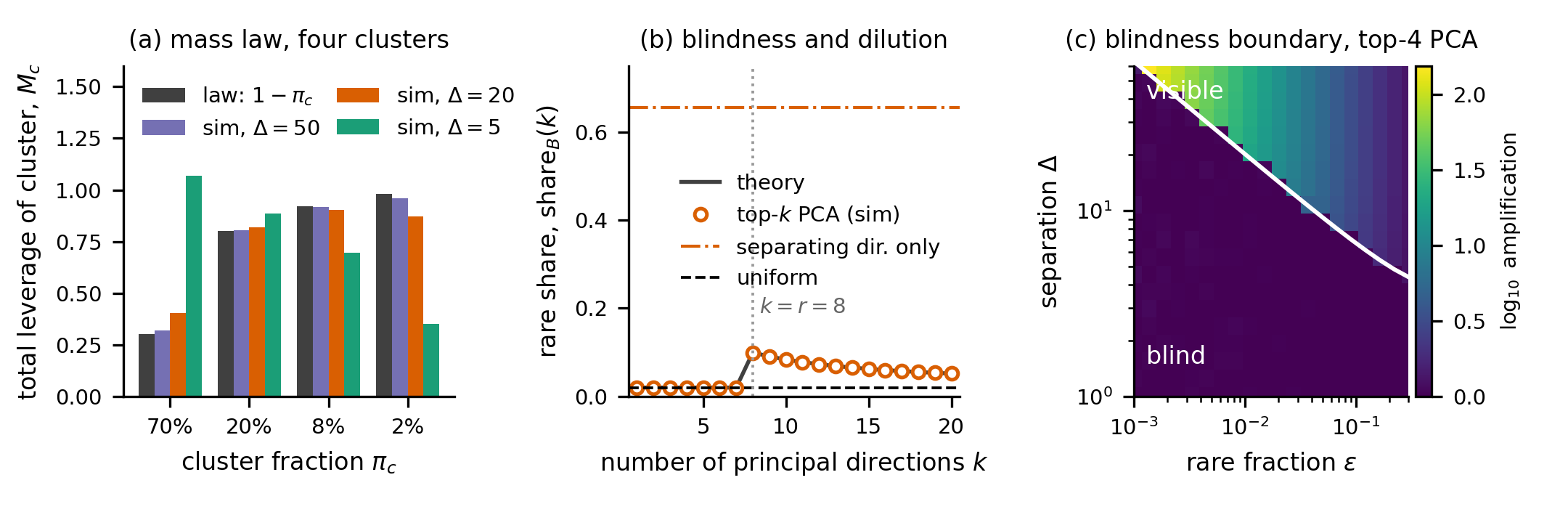}
\caption{Leverage behavior in higher-dimensional buffers. (a) Multi-cluster leverage mass \(M_c\) versus the law \(1-\pi_c\) in (9). (b) Rare share under top-\(k\) PCA leverage when the separating direction has variance rank \(r=8\); lines show (10) and leverage along the separating direction alone. (c) Rare-cluster amplification under top-4 PCA leverage over \((\epsilon,\Delta)\); white curve shows the predicted boundary (11)}
\label{fig:theory2}
\end{figure*}

Three consequences of this analysis inform our method. \emph{(i) Amplification is separation-limited, not rarity-limited.} Making the event rarer at fixed $\Delta$ gives $\shB/\varepsilon\to(\Delta^{2}+s_B^{2})/s_A^{2}$, so $p_B(\beta)/\varepsilon\to1+\beta\big[(\Delta^{2}+s_B^{2})/s_A^{2}-1\big]$, which for equal widths is $1+\beta\Delta^{2}/s^{2}$; at $\beta=\tfrac12$ the amplification saturates at $1+\Delta^{2}/2s^{2}$, i.e.\ $13.5\times$, $51\times$, and $201\times$ for $\Delta=5$, $10$, $20$, not at the naive $1/2\varepsilon$ that would put half of every batch in the rare cluster ($250\times$ at $\varepsilon=0.002$). The ceiling is reached to within $10\%$ once $\varepsilon\Delta^{2}/s_A^{2}\lesssim0.1$: at $\varepsilon=0.002$ the simulated amplifications are $12.9\times$, $42.8\times$, and $111\times$, so the $\Delta{=}5$ curve is at its ceiling while the $\Delta{=}20$ curve is still far below it (it would need $\varepsilon\lesssim3\times10^{-4}$). In practice, a rare mode that is not geometrically distinct from routine operation will not be amplified much no matter how rare or how compact it is, and the boost of one that is distinct can be predicted from $\Delta/s_A$. \emph{(ii) Batch composition is exactly linear in $\beta$}, \eqref{eq:batchshare}, and a stratified batch made of $(1-\beta)B$ uniform draws and $\beta B$ leverage draws has the same expected composition as $B$ draws from the mixture (Fig.~\ref{fig:theory}b, agreement within $0.002$); an implementation can therefore keep its uniform sampler and simply append a leverage-sampled half-batch. \emph{(iii) Multi-cluster mass law.} With $q$ clusters of fractions $\pi_1,\dots,\pi_q$ whose centers are far apart compared with their spreads, the centered centers span a $(q{-}1)$-dimensional subspace; taking $\mathcal{K}$ to be that subspace, every point of cluster $c$ sits essentially at its center, and \eqref{eq:leverage} gives it the score $1/(N\pi_c)-1/N$ (the $N\pi_c$ points of a cluster share one location's pull on the fit, minus a centering correction), the total leverage cluster $c$ mass is
\begin{equation}
M_c=1-\pi_c,\qquad \textstyle\sum_c M_c=q-1=|\mathcal{K}| ,
\label{eq:masslaw}
\end{equation}
and pure-leverage shares are $(1-\pi_c)/(q-1)$: a cluster's share depends only on its size, with an amplification $(1-\pi_c)/[(q-1)\pi_c]$ that exceeds one for small clusters and falls below one for large ones. For four clusters of fractions $0.70/0.20/0.08/0.02$, each minority cluster displaced by $\Delta$ along its own feature column, the $2\%$ cluster is amplified $14.5\times$ and the $70\%$ cluster suppressed to $0.19\times$ at $\Delta=20$, approaching the limits $16.3\times$ and $0.14\times$ ($16.0\times$ and $0.15\times$ at $\Delta=50$; Fig.~\ref{fig:theory2}a); the law requires $\Delta^{2}\gg1/\pi_c$ for every cluster. The three highest-kurtosis columns are exactly the three displaced columns, so the selection rule recovers this subspace with no labels anywhere.

Real buffers vary along many directions with very different variances, and the rare cluster is displaced along only one of them, the \emph{separating direction}, whose total variance is $\lambda=\bar s^{2}+\varepsilon(1-\varepsilon)\Delta^{2}$ from \eqref{eq:lambda}: its within-cluster variance plus the small bump. Let $r$ be its \emph{variance rank}, its position when all directions of the buffer are sorted from largest to smallest variance. If $\mathcal{K}$ is the top-$k$ principle directions, then for $k<r$ the separating direction is not in the subspace at all and leverage is \emph{blind}, $\shB=\varepsilon$; for $k\ge r$ it is included together with $k-1$ background directions that carry no information about the clusters, and since every transition, rare or common, receives on average one unit of score along each such direction, the share is \emph{diluted} to
\begin{equation}
\shB(k)=\frac{\varepsilon\,(k-1+m_B)}{\varepsilon\,(k-1+m_B)+(1-\varepsilon)(k-1+m_A)},
\label{eq:dilution}
\end{equation}
with $m_B,m_A$ from \eqref{eq:mAmB}; this equals \eqref{eq:share} at $k=1$ and decays back toward $\varepsilon$ as $k$ grows. In the example of Fig.~\ref{fig:theory2}b ($r=8$), top-$k$ leverage amplifies the rare cluster $1.0\times$ for $k<8$, $5.0\times$ at $k=8$, and $2.6\times$ with all $d=20$ directions (the full Mahalanobis score), against $32.8\times$ for the separating direction alone. For the boundary between the two regimes, the separating direction enters the top $k$ when the bump lifts its variance above the $k$-th largest background variance $\lambda^{\mathrm{w}}_{(k)}$,
\begin{equation}
\varepsilon(1-\varepsilon)\Delta^{2}\;>\;\lambda^{\mathrm{w}}_{(k)}/\bar s^{2}-1 ,
\label{eq:boundary}
\end{equation}
which reproduces the blind/visible boundary of the simulated $(\varepsilon,\Delta)$ phase diagram in $97\%$ of its cells (Fig.~\ref{fig:theory2}c). Ridge leverage, $\ell_i^{\gamma}=\tfrac{1}{N-1}\bar x_i^{\top}(C+\gamma I)^{-1}\bar x_i$, keeps every direction and shrinks the low-variance ones, so it is never blind, but it is diluted by the effective dimension $d_{\mathrm{eff}}(\gamma)=\sum_j\lambda_j/(\lambda_j+\gamma)$ that it retains; its optimum is interior (too small a $\gamma$ dilutes over all $d$ directions, too large a $\gamma$ shrinks away a low-variance separating direction), and in the same example the best $\gamma$ recovers only $2.9\times$. The design goal is thus the \emph{smallest} score subspace containing the separating directions. Variance ranking does not target it, because it orders directions by how much the data spreads, not by whether a small, far-away group is present; a column carrying such a group has heavy tails regardless of its variance, which is the motivation for tail-based column selection of Sec.~\ref{sec:columns}.

\section{EXPERIMENTS}
\label{sec:experiments}

\subsection{Simple 2D Grid World}
As a proof-of-principal experiment we consider a Q-learning Deep RL agent trying to find the goal in a simple 2D grid world which is a $24\times 12$ lattice ($x\in \{0,\dots,23\}$, $y\in \{0,\dots,11\})$, with a main room consisting of 144 cells located at $x\in\{0,11\}$, $y\in\{0,11\}$, a narrow corridor whose cells are $x\in\{12,\dots,15\}$, $y=5$, and an annex whose cells are $x\in\{16,\dots,23\}$, $y\in\{2,\dots,9\}$. The goal in this case is always located at position $(1,5)$ in the main room. The goal is in the main room, escaping the annex requires values propagated along the chain: annex $\rightarrow$ corridor $\rightarrow$ room, this structure is invisible when focused on only main-room data. We model this as a Markov decision process (MDP) tuple $(\mathcal{S},\mathcal{A},P,r,\gamma)$, a state space, action space, transition probabilities $P(\mathbf{s}'|\mathbf{s},a)$, a reward function, and a discount factor. Here the state is the cell we occupy, $\mathbf{s}=(x,y)$; the actions are $\mathcal{A}=$(up, down, left, right); the transition rule is ``move one cell in the chosen direction, except with probability $0.1$ the action {\it{slips}} to a uniformly random one, and bumping into a wall leaves you in place"; the reward is $r=-1$ for every step; the episode ends at the goal $(1,5)$; and $\gamma=0.99$. Maximizing the return is equivalent to minimizing the number of steps required to reach the goal. Starting at $t=0$ and taking $T$ steps to reach the goal gives
\begin{equation}
    G = \sum_{t=0}^{T-1}\gamma^t(-1) = - \frac{1-\gamma^T}{1-\gamma}. \label{T_step_cost}
\end{equation}
For a policy $\pi$, if action $a$ is taken now and $\pi$ is followed afterwards, the expected return is
\begin{equation}
    Q^\pi(\mathbf{s},a) = \mathbb{E}\left[G_t | \mathbf{s}_t = \mathbf{s}, a_t=a, \text{ then follow } \pi \right ],
\end{equation}
and the optimal action-value function is $Q^*(\mathbf{s},a)=\max_\pi Q^\pi(\mathbf{s},a)$, which is the best expected return achievable from $(\mathbf{s},a)$, whose state version is $V^*(\mathbf{s})=\max_aQ^*(\mathbf{s},a)$. According to the Bellman optimality equation, $Q^*$ is the optimal solution of
\begin{eqnarray}
    Q^*(\mathbf{s},a) &=& \mathbb{E}_{\mathbf{s}'\sim P(\cdot|\mathbf{s},a)} \left[ r + \gamma \max_{a'} Q^*(\mathbf{s}',a') \right ] \\
    &=&-1 + \gamma \sum_{a''} p(a''|a)V^*(\text{next}(\mathbf{s},a'')),
\end{eqnarray}
where $V^*(\text{goal})=0$ and we have simplified the equation because in our case $r \equiv -1$, and the randomness comes from the slip so $a''$ represents the action actually taken. We can use this rule iteratively to find $Q^*$ exactly for this simple finite-state grid world.  The setup is shown in Figure \ref{fig:grid_world} with the image on the far right colored by $Q^*$ for each cell. For our Q-learning setup, we assume we do not have access to $P$ and we use a fixed replay buffer of transitions $(\mathbf{s},a,r,\mathbf{s}')$, where Q-learning replaces the expectation in the Bellman equation with a single sampled transition, each giving a temporal-difference (TD) target
\begin{equation}
    y = r + \gamma(1-\text{done})\max_{a'}Q(\mathbf{s}',a'),
\end{equation}
and we train a neural network model to learn $Q_\theta(\mathbf{s},a)$ by minimizing the training cost function $(Q_\theta(\mathbf{s},a)-y)^2$ over mini-batches from the buffer. As $Q_\theta(\mathbf{s},a)$ is trained, for each state $\mathbf{s}=(x,y)$ it outputs four numbers which are its estimates of the expected discounted return for each action taken from that cell. TThe agent then chooses actions greedily according to
\begin{equation}
    \pi(\mathbf{s}) = \operatorname*{argmax}_{a\in\mathcal{A}}Q_\theta(\mathbf{s},a). \label{greedy_policy}
\end{equation}
We perform an experiment in which we fill a buffer with $N=40,000$ transitions $(\mathbf{s},a,r,\mathbf{s}')$, with a fraction $\epsilon=0.05$ of them coming from the rare corridor + annex region and $1-\epsilon=0.95$ from the main room. We use the approach described above, standardizing every column by subtracting its mean and dividing by its standard deviation, and consider the excess kurtosis of each standardized column
\begin{equation}
    \bar{x}_{ij} = \frac{x_{ij}-\mu_j}{c_j}, \quad \kappa_j = \frac{1}{N}\sum_{i}\bar{x}^4_{ij}-3.
\end{equation}
In this simple example, we score only the state coordinates because $r\equiv-1$, and get excess kurtosis values of $1.87$ for the $x-$column and $-1.19$ for the $y$-column. We then compare various buffer sampling approaches for training. 

\begin{figure}[t]
\centering
\includegraphics[width=\columnwidth]{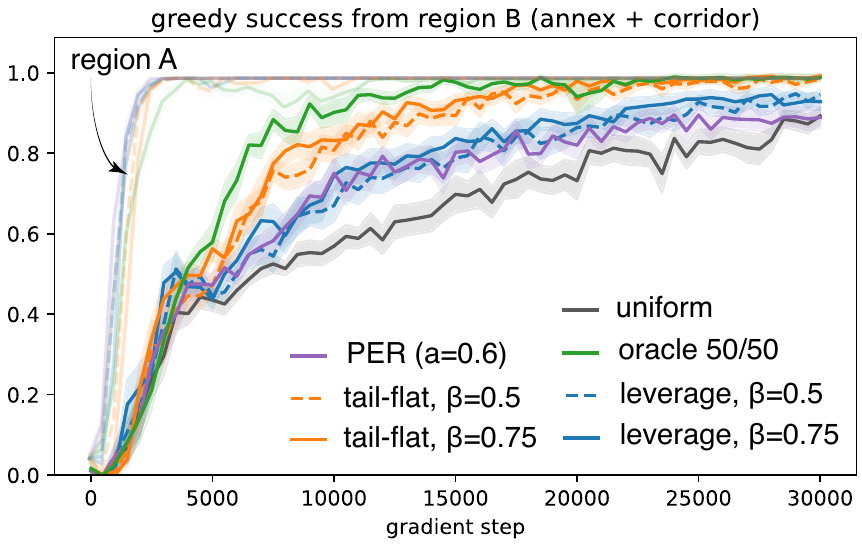}
\caption{Success rate for each method for regions A and B as a function of training steps of $Q_\theta(\mathbf{s},a)$ are shown for the various methods. All methods quickly succeed in the common region A (opaque curves), and their performance separates in the rare/difficult region B with our proposed methods all outperforming PER and uniform sampling.}
\label{fig:grid_world_results}
\end{figure}

\begin{itemize}
    \item \textbf{Uniform:} $p_i=1/N$ to each entry in the buffer.
    \item \textbf{$\boldsymbol{\beta}$-leverage:} 
    \begin{equation}
        p_i = (1-\beta)\frac{1}{N} + \beta \frac{\ell_i}{\sum_j \ell_j}, \quad \beta\in[0,1]. \quad \ell_j=z^2_{ij^*} \label{eq:mixture}
    \end{equation}
    Column of standardized $x$ values chosen by kurtosis.
    \item \textbf{Tail-flat:} This approach uses the score to only find the top-decile tail $T$ and then samples uniformly inside of there, taking $T$ as the top-decile leverage tail, giving
    \begin{equation}
        p_i = (1-\beta)\frac{1}{N} + \beta \frac{\mathbf{1}[i \in T]}{|T|}.
    \end{equation} \label{eq:tail_flat}
    \item \textbf{Oracle:} The oracle knows rare/common locations uniformly samples half of its values from each.
\end{itemize}
We use a total batch size of $B=256$ and our network $Q_\theta:\mathbb{R}^2\rightarrow\mathbb{R}^4$ is a simple dense network with 2 hidden layers of width 24 each and ReLU activation functions. Once we start training the network, we check its performance on the common room region (A) and on the rare corridor + annex region (B) after every 500 training (network weight gradient update) steps. For each check, we start in every single one of the 68 cells in region B as well as in every single one of the 143 cells in region A, apply the greedy policy of Equation (\ref{greedy_policy}) for a maximum of 100 steps, and record the total fraction of starts which reach the goal within those 100 steps. We also calculate the root mean squared error of $Q_\theta$ in each region $\mathcal{R} \in \{\text{A}, \text{B} \}$, relative to the ground truth $Q^*$ according to
\begin{equation}
    \text{RMSE}_{\mathcal{R}} = \sqrt{\frac{1}{4|\mathcal{R}|}\sum_{\mathbf{s}\in\mathcal{R}}\sum_{a} \left [Q_\theta(\mathbf{s},a) - Q^*(\mathbf{s},a) \right ]^2 }.
\end{equation}
We repeat this for 20 different random seeds, for $\beta=0.5$ and $\beta=0.75$. The success rate vs training step results are shown in Figure \ref{fig:grid_world_results} and are summarized in Table \ref{tab:sampling_conditions}.

\begin{table}[]
\caption{Success rate on task B and Q-value RMSE on tasks B and A under each sampling condition (mean $\pm$ std.) over 20 seeds.}
\label{tab:sampling_conditions}
\centering
\begin{tabular}{@{}lccc@{}}
\toprule
 & & \multicolumn{2}{c}{Q-RMSE} \\
\cmidrule(lr){3-4}
Condition & Final succ.\ B & B & A \\
\midrule
Uniform               & $0.882 \pm 0.018$ & 3.33 & 0.31 \\
Leverage ($\beta=0.5$)    & $0.941 \pm 0.010$ & 2.42 & 0.36 \\
Leverage ($\beta=0.75$)   & $0.928 \pm 0.021$ & 1.43 & 0.38 \\
Tail-flat ($\beta=0.5$)   & $0.978 \pm 0.007$ & 1.63 & 0.57 \\
Tail-flat ($\beta=0.75$)  & $\textbf{0.989} \pm \textbf{0.006}$ & 1.58 & 0.87 \\
PER ($\alpha=0.6$)    & $0.888 \pm 0.024$ & 2.71 & 0.26 \\
Oracle 50/50          & $0.987 \pm 0.009$ & 1.06 & 0.72 \\
\bottomrule
\end{tabular}
\end{table}

\begin{table}[t]
\caption{TD3+BC for 20 seeds \& 10k gradient steps. Orbit performance is averaged over the last three evaluations.}
\label{tab:td3bc_sampling}
\centering
\small
\begin{tabular}{@{}lccc@{}}
\toprule
 & Orbit & & Batch \\
Condition & (last-3) & Straight & orbit share \\
Uniform                     & 0.49 & 1.00 & 0.069 \\
Window tail-flat ($b=0.5$)  & \textbf{0.69} & 1.00 & 0.220 \\
Leverage ($\beta=0.5$)          & 0.51 & 1.00 & 0.110 \\
Leverage ($\beta=0.75$)         & 0.44 & 1.00 & 0.131 \\
Tail-flat ($\beta=0.5$)         & 0.45 & 1.00 & 0.098 \\
Tail-flat ($\beta=0.75$)        & 0.38 & 0.99 & 0.113 \\
PER ($\alpha=0.6$)          & 0.51 & 0.99 & 0.084 \\
Oracle 50/50                & 0.74 & 1.00 & 0.500 \\
\bottomrule
\end{tabular}
\end{table}

\begin{figure}[]
\centering
\includegraphics[width=\columnwidth]{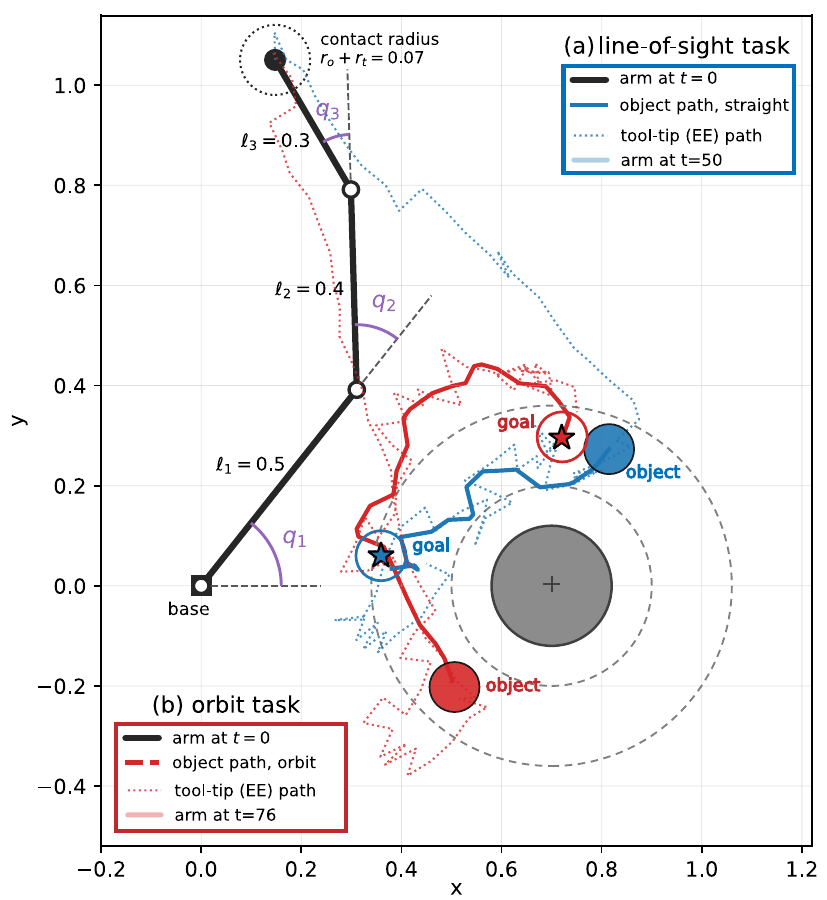}
\caption{The dashed circles show the annulus from which the object start $o_0$ and goal $g$ are drawn. (a) For straight tasks the line of sight $o_0\!\to\! g$ clears the obstacle. (b) For orbit tasks the goal lies behind the obstacle and the object must be pushed around it. Solid curves are logged object paths (scripted controller with exploration noise), dotted curves the tool-tip path.}
\label{fig:robot_arm}
\end{figure}

\begin{figure}[]
\centering
\includegraphics[width=\columnwidth]{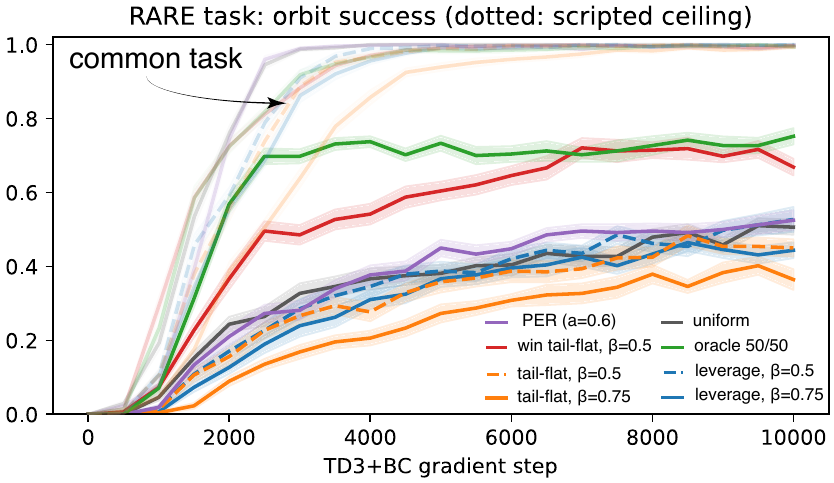}
\caption{Comparison of the various approaches for the 3-joint robot arm.}
\label{fig:robot_arm_results}
\end{figure}

\subsection{Robot Arm \& Window Tail-Flat}

The gridworld's rare region B is  the set of large $x$, so the kurtosis rule found it in a single column. The second experiment is built so that this is not the case. A three-link planar arm (Fig. \ref{fig:robot_arm}) pushes a disk from a start $o_0$ to a goal $g$ past a fixed obstacle centered at $C$; the log is generated by a scripted controller with exploration noise, with $950$ \emph{straight} episodes, in which the line of sight from $o_0$ to $g$ clears the obstacle, and $35$ \emph{orbit} episodes, in which $g$ lies behind it and the object must be pushed around; the learner is TD3+BC~\cite{fujimoto2021minimalist}, trained offline on this fixed log. The object start $o_0$ and goal $g$ are drawn from the same annulus around $C$, so every object and goal coordinate has the same marginal; what makes an orbit transition rare is where the goal sits relative to the object and the obstacle, and that this relation persists for tens of steps while the object travels around. Per-transition scoring of the state columns does not see this, so the rule of Sec.~III-B falls back on a joint-angle tail. Extreme arm configurations arereached in both families, and the row-level leverage and tail-flat samplers raise the orbit share of a batch only from $0.069$ to $0.10$--$0.13$, leaving orbit success at the uniform level. This is the failure mode anticipated in Sec.~III-B, where the choice of which columns enter $x_i$ was flagged as decisive.

\textbf{Window tail-flat.} Each logged episode is cut into consecutive windows $w$ of $L_w=24$ transitions, and a window is described relative to its own goal,
\begin{equation}
  z_w=\big[\,g_w,\ \ g_w-\bar{o}_w,\ \ \lVert g_w-\bar{o}_w\rVert\,\big]\in\mathbb{R}^{5},
  \label{eq:win-desc}
\end{equation}
where $\bar{o}_w=\frac{1}{|w|}\sum_{i\in w}o_i$ is the mean object position over the window and $g_w$ its (episode-constant) goal: where the goal is, the offset from where the object currently is to it, and how far it still has to go. The descriptors are standardized as in (\ref{eq:standardize}) and scored by the selection-free Mahalanobis form (\ref{eq:mahal}) over all $d_w=5$ descriptor directions,
\begin{equation}
  \ell_w=\bar{z}_w^{\top}R_z^{-1}\bar{z}_w ,
  \label{eq:win-score}
\end{equation}
with $R_z$ the descriptor correlation matrix; with $\approx N/L_w$ windows and $d_w=5$ this is one $5\times5$ inverse. The top-decile tail $T_w=\{w:\ell_w\ge Q_{0.9}(\ell)\}$ is then replayed with the tail-flat randomly chosen as in (\ref{eq:tail_flat}) applied to windows with $p=0.5$ a window uniformly and a transition uniformly inside it; $p=0.5$, a uniform transition
\begin{equation}
  p_i=(1-\beta)\frac{1}{N}+\beta\,\frac{\mathbf{1}[w(i)\in T_w]}{|T_w|\,|w(i)|},
  \qquad \beta=0.5 ,
  \label{eq:win-tailflat}
\end{equation}
where $w(i)$ is the window containing transition $i$. As in (8) the orbit share of a batch is linear in $\beta$, $(1-\beta)\varepsilon+\beta\,\pi_T$ with $\pi_T$ the orbit fraction of the tail's transitions; here $\pi_T\approx0.37$, more than five times the base rate $\varepsilon=0.069$, so at $\beta=0.5$ the batch share rises to $0.22$ with no labels, rewards, or critic involved. Window tail-flat is thus the rule of Sec. III with two of its free choices made differently. The unit scored is a window rather than a transition, and the feature row is a goal-relative summary of that window rather than raw state columns. The score (\ref{eq:mahal}), the tail, and the mixture (\ref{eq:tail_flat}) are unchanged.  Table \ref{tab:td3bc_sampling} and Figure \ref{fig:robot_arm_results} summarize the results.

\section{Conclusion}

We introduced a diversity-aware experience replay framework that uses statistical leverage to preferentially revisit underrepresented experience without relying on rewards, labels, or the current value function. Analysis of Gaussian-mixture buffers characterized when leverage amplifies rare data, showing that amplification is governed by geometric separation rather than rarity alone. The analysis also revealed that variance-ranked subspaces can miss rare modes, motivating kurtosis-based feature selection. In the gridworld, the resulting leverage and tail-based replay rules improved learning in the underrepresented region compared with uniform replay and PER. The robot-arm experiment further showed that the representation used for scoring is critical: when rare behavior could not be distinguished from individual state coordinates, transition-level scoring provided little improvement, whereas scoring short goal-relative trajectory windows increased rare-task success from 0.49 to 0.69, compared with 0.74 for a labeled oracle. These results suggest that statistical leverage provides a simple, critic-independent mechanism for rebalancing imbalanced replay buffers.


\bibliographystyle{IEEEtran}
\bibliography{references_V2}

\end{document}